\documentclass[sigconf]{acmart}

\usepackage{graphicx}
\usepackage{subfigure}
\usepackage{algorithm}
\usepackage{algorithmic}
\usepackage{multirow}

\def\BibTeX{{\rm B\kern-.05em{\sc i\kern-.025em b}\kern-.08emT\kern-.1667em\lower.7ex\hbox{E}\kern-.125emX}}
    
\copyrightyear{2019}
\acmYear{2019}
\setcopyright{acmlicensed}
\acmPrice{15.00}
\acmDOI{10.1145/1122445.1122456}
\acmISBN{978-1-4503-9999-9/18/06}

\begin{document}

%
\title{Learning Cross-Domain Representation with Multi-Graph Neural Network}
%

%

\author{Yi Ouyang}
\authornote{First author.}
\authornote{This work was done when Yi Ouyang worked as intern at Tencent.}
\affiliation{%
	\institution{Northwestern Polytechnical University}
}

\author{Bin Guo}
\authornote{Corresponding author. E-mail: guobin.keio@gmail.com}
\affiliation{%
	\institution{Northwestern Polytechnical University}
}

\author{Xing Tang}
\authornote{Co-first author.}
\affiliation{%
	\institution{Tencent}
}

\author{Xiuqiang He}
\affiliation{%
	\institution{Tencent}
}

\author{Jian Xiong}
\affiliation{%
	\institution{Tencent}
}

\author{Zhiwen Yu}
\affiliation{%
	\institution{Northwestern Polytechnical University}
}

\begin{abstract}
Learning effective embedding has been proved to be useful in many real-world problems, such as recommender systems, search ranking and online advertisement. However, one of the challenges is data sparsity in learning large-scale item embedding, as users' historical behavior data are usually lacking or insufficient in an individual domain. In fact, user's behaviors from different domains regarding the same items are usually relevant. Therefore, we can learn complete user behaviors to alleviate the sparsity using complementary information from correlated domains. It is intuitive to model users' behaviors using graph, and graph neural networks (GNNs) have recently shown the great power for representation learning, which can be used to learn item embedding. However, it is challenging to transfer the information across domains and learn cross-domain representation using the existing GNNs. To address these challenges, in this paper, we propose a novel model - \textbf{D}eep \textbf{M}ulti-\textbf{G}raph \textbf{E}mbedding (DMGE) to learn cross-domain representation. Specifically, we first construct a multi-graph based on users' behaviors from different domains, and then propose a multi-graph neural network to learn cross-domain representation in an unsupervised manner. Particularly, we present a multiple-gradient descent optimizer for efficiently training the model. We evaluate our approach on various large-scale real-world datasets, and the experimental results show that DMGE outperforms other state-of-art embedding methods in various tasks.
\end{abstract}


%

%
\maketitle

\section{Introduction}
Recently, many online personalized services have utilized users' historical behavior data to characterize user preferences, such as: online video sites~\cite{covington2016deep}, App stores~\cite{cheng2016wide}, online advertisements~\cite{guo2017deepfm} and E-commmerce sites~\cite{zhao2018learning,wang2018billion}. Learning the representation from user-item interactions is an essential issue in most personalized services. Usually, low-dimensional embeddings can effectively represent attributes of items and preferences of users in a uniform latent semantic space, which are helpful to provide personalized services and improve user experience. Moreover, the representation of users and items has been widely applied to many research topics related to above real-world scenarios, including: large-scale recommendation~\cite{wang2018billion,ying2018graph}, search ranking~\cite{grbovic2018real,chu2018deep}, cold-start problem~\cite{zhao2018learning}.


In large-scale personalized services, there are usually a relative small portion of active users, and a majority of non-active users often interact with only a small number of items, users' behavior data is thus lacking or insufficient in an individual domain, which makes it difficult to learn effective embeddings~\cite{wang2019solving}. On the other hand, though data from a single domain is sparse, users' behaviors from correlated domains regarding the same items are usually complementary~\cite{zhuang2017sequential}. Take the App store as an example, there are two ways users interact (e.g., download) with items (i.e., Apps). One is downloading Apps recommended on the homepage or category pages of App store (i.e., \emph{recommendation domain}), the other is by searching (i.e., \emph{search domain}). User behaviors in the search domain reflect user's current needs or intention, while that in the recommendation domain represent user's relative long-term interests. Leveraging the interaction data from the search domain can improve the performance of recommendation. On the other hand, interaction data from the recommendation domain can also help to explore user's personalized interests and therefore optimize the ranking list in search domain. Therefore, we are motivated to leverage the complementary information from correlated domains to alleviate the sparsity problem.

\begin{figure}
	\centering
	\subfigure[Users' behavior sequences in multiple domains.]{
		\begin{minipage}{0.45\textwidth}
			\centering
			\includegraphics[height=38mm]{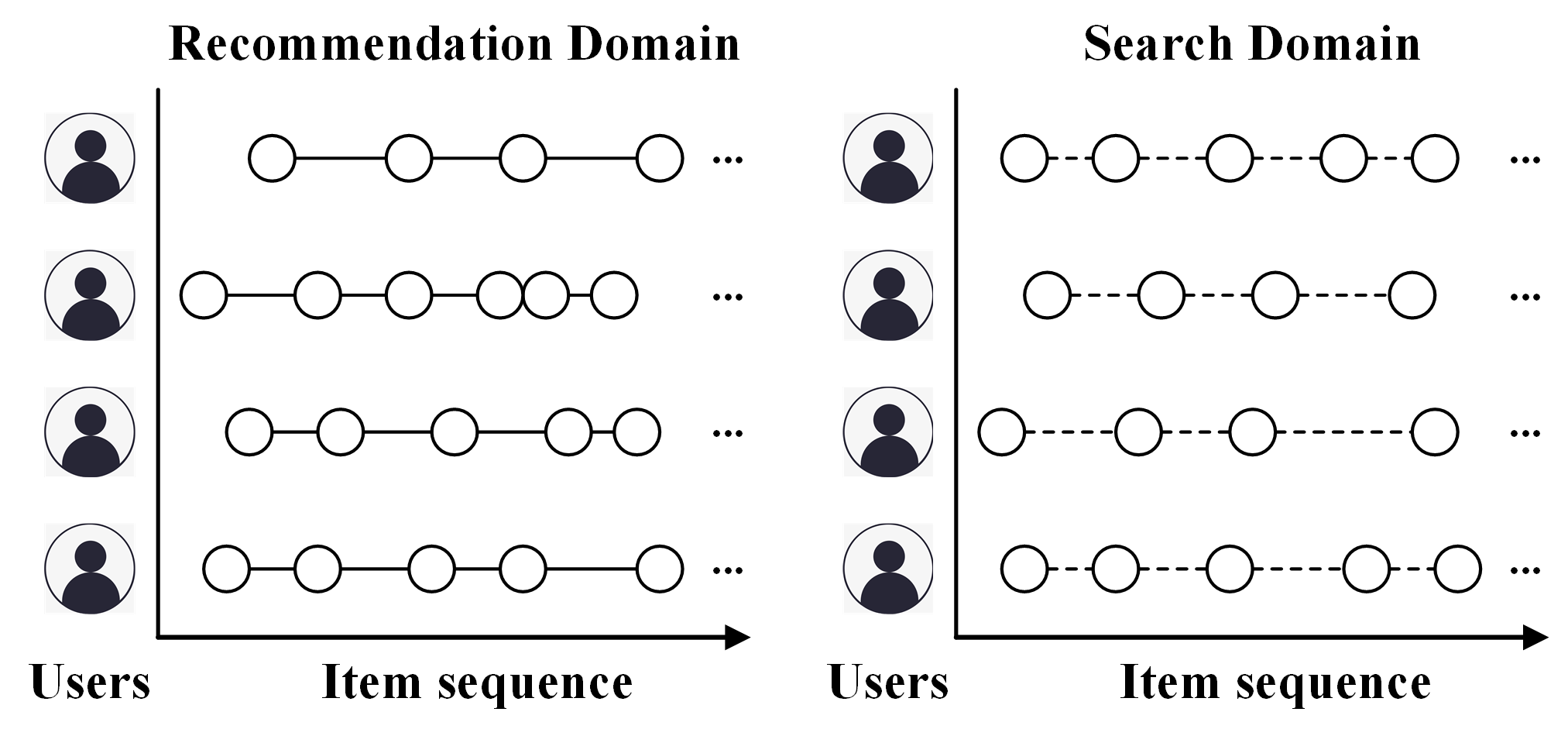}
	\end{minipage}}
	\subfigure[Item graphs.]{
		\begin{minipage}{0.26\textwidth}
			\centering
			\includegraphics[height=22mm]{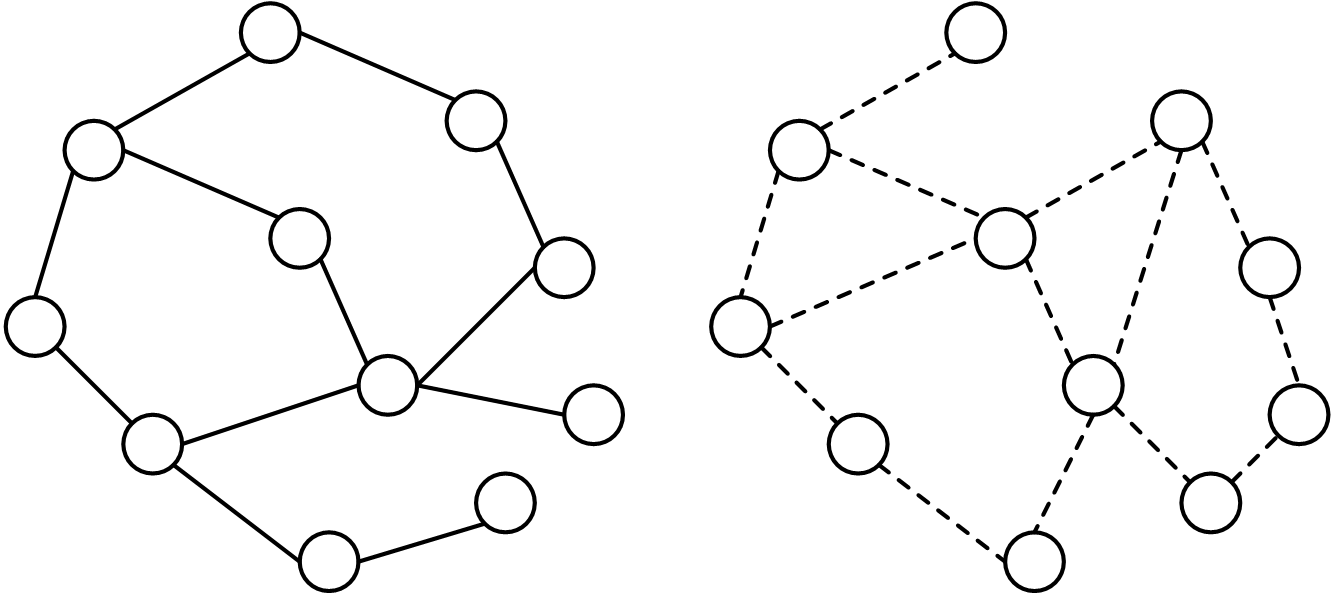}
	\end{minipage}}
	\subfigure[Multi-graph.]{
		\begin{minipage}{0.19\textwidth}
			\centering
			\includegraphics[height=22mm]{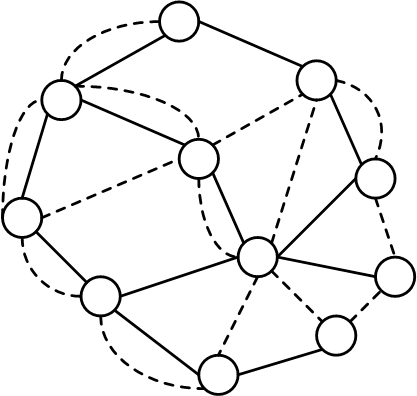}
	\end{minipage}}
	\caption{The construction of multi-graph.}
	\label{fig:multi_graph}
\end{figure}

Generally, users' behaviors are sequential~\cite{hidasi2016session} (take the App store as example, as shown in Figure~\ref{fig:multi_graph} (a)), and graph can be used to model users' sequential behaviors intuitively~\cite{wang2018billion}. Specifically, in each domain (as shown in Figure~\ref{fig:multi_graph} (b)), we can construct an item graph by modeling the items as nodes, the item co-occurrences as edges, and the number of co-occurrences in all users' behavior sequences as the weights of edges. Through applying graph embedding methods such as DeepWalk~\cite{perozzi2014deepwalk,wang2018billion}, it can generate abundant item sequences by running random walk on the item graph, and then use the Skip-Gram algorithm~\cite{mikolov2013efficient,mikolov2013distributed} to learn item embedding. Compared with the random walk based graph embedding methods~\cite{perozzi2014deepwalk,grover2016node2vec}, graph neural networks (GNNs) have shown the great power for representation learning on graphs recently~\cite{xu2019powerful}. As a state-of-the-art GNN, graph convolutional network (GCN)~\cite{kipf2017semi} is proposed based on convolutional neural networks, and generates node embedding by operating convolution on the graph. The graph convolution operation in GCN is to encode node attributes and graph structure using neural networks, thus GCN performs well in graph embedding, and can be used for item embedding. However, these methods are developed for learning single graph embedding, i.e., single domain embedding. Users' behaviors in cross-domain are more complex, and it is more reasonable to model user's behaviors as multi-graph (as shown in Figure~\ref{fig:multi_graph} (c)), which consists a set of nodes and multiple types of edges (i.e., solid and dashed lines represent two types of edges). Concretely, nodes represent the same items across domains and each type of edge denotes the co-occurrences of item pairs in each domain. In multi-graph, there may exist multiple types of edges between pairs of nodes, each type of edge forms a certain subgraph (i.e., a domain), and these subgraphs are related to each other, as all of them share the same nodes. Thus, it is likely that each node (i.e. item) in different subgraph (i.e. domain) has a different representation, and all these representations of a node are relevant to each other. 

However, the existing single graph embedding methods fail to fuse the complex relations in multi-graph, and generate effective node embeddings. The cross-domain scenario poses challenges to transfer the information across domains and learn cross-domain representation. On the other hand, though GCN is effective, stacking many convolutional layers makes GCN difficult to train, as the iterative graph convolution operation is prone to overfit, as stated in~\cite{li2018deeper}. It brings additional complexity and challenges to apply GCN to learn cross-domain (or multi-graph) representation. Thus, to better utilize the power of GCN, dedicated efforts are desired to design a novel neural network architecture based on GCN for cross-domain representation learning, and optimize the neural network efficiently to overcome the disadvantages of GCN.

To address these challenges, in this paper, we propose a novel embedding model, named Deep Multi-Graph Embedding (DMGE). We first construct the item graph as a multi-graph based on users' sequential behaviors from different domains. Specifically, the nodes in multi-graph represent items, and two nodes are connected by an edge if they consecutively occur in one user's sequence. Thus, learning the item embedding is converted to learn node embedding in the multi-graph. To utilize the power of GCN on graph embedding, we propose a graph neural network inspired by multi-task learning regime, which extends GCN to learn cross-domain representation. Specifically, each domain is viewed as a task in the model, and we design the domain-specific layers to generate domain-specific representation for each domain, all domains are correlated by the domain-shared layers, which generate domain-shared representation. The model is then trained in an unsupervised manner by learning the graph structure. Besides, to overcome the disadvantages of GCN, we introduce a multiple-gradient descent optimizer to train the proposed model, which can adaptively adjust the weight of each domain. Particularly, It updates the parameters of the domain-shared layers by using the weighted summation of the gradients of all domains, and parameters of the domain-specific layers by using the gradients of the specific domain. The main contributions of our work are summarized as follows:



\begin{itemize}
	\item We focus on learning cross-domain representation. Innovatively, we model users' behaviors in cross-domain as multi-graph, and propose a graph neural network to learn domain-shared and domain-specific representation, simultaneously.
	
	\item  We propose a novel embedding model named Deep Multi-Graph Embedding (DMGE), which is a graph neural network based on multi-task learning. Particularly, we present a multiple-gradient descent optimizer to efficiently train the model in an unsupervised manner.
	
	\item We evaluate DMGE on various large-scale real-world datasets, and the experimental results show that DMGE outperforms other state-of-the-art embedding methods in various tasks.
\end{itemize}

\section{Deep Multi-Graph Embedding}

In this section, we elaborate the Deep Multi-Graph Embedding (DMGE) model for cross-domain item embedding. We first present the problem definition. Then, we propose a multi-graph neural network to learn node embedding in the multi-graph. Finally, we present an multiple-gradient descent optimizer to efficiently train the model in an unsupervised manner.

\subsection{Problem Definition}
Suppose there are $D$ domains. For each domain $d$ $(d=\{1, ..., D\})$, we first construct the item graph as an undirected weighted graph $\mathcal{G}_d=(\mathcal{V}, \mathcal{E}_d)$. As these $D$ domains are correlated and share the same set of items, we then construct the cross-domain item graph as an undirected weighted multi-graph $\mathcal{G}=(\mathcal{V}, \mathcal{E})$, which contains the node set $\mathcal{V}$ with $N$ nodes and the edge set $\mathcal{E}$ with $D$ types of edges, i.e., $\mathcal{E} = \{\mathcal{E}_1, ..., \mathcal{E}_D\}$.

Our problem can be formally stated as follows, with an undirected weighted multi-graph $\mathcal{G}=(\mathcal{V}, \mathcal{E})$, and the node feature matrix $\mathbf{X} \in \mathbb{R}^{N \times M}$, representing input for each node as an $M$-dimensional feature vector, our goal is to learn a set of embedding for all nodes in each subgraph $\mathcal{G}_d$, i.e., $\mathcal{X} = \{\mathbf{X}_1, ..., \mathbf{X}_D\}$ $($$\mathbf{X}_d \in \mathbb{R}^{N \times E}$  is the node embedding in subgraph $\mathcal{G}_d$, with each node has an $E$-dimensional embedding$)$ by solving the following optimization problem:

\begin{equation}
\min
\left\{
\begin{aligned}
& \sum_{v_i \in \mathcal{V}} \sum_{v_j \in N(v_i, 1)} -\log p(v_j|v_i, 1)\\
& \, \, \,  ...\\
& \sum_{v_i \in \mathcal{V}} \sum_{v_j \in N(v_i, D)} -\log p(v_j|v_i, D)
\end{aligned}	
\right.
\label{eq:problem}
\end{equation}
where $p(v_j|v_i, d)$ is the probability that there exists an edge between node $v_i$ and node $v_j$ in the subgraph $\mathcal{G}_d$, and $N(v_i, d)$ is the set of neighborhood of node $v_i$ in the subgraph $\mathcal{G}_d$.

\subsection{Multi-Graph Neural Network}
\label{sec:MGNN}

As is discussed, graph neural networks (GNNs)~\cite{zhou2018graph,xu2019powerful} have emerged as a powerful approach for representation learning on graphs recently, such as GCN~\cite{kipf2017semi}. Thus we emphasize on applying GCN for multi-graph embedding. In a multi-graph, the same set of nodes are shared in all subgraphs. For each node, it has different neighbors in different subgraph, thus it is likely that it has different representation in different subgraph. Moreover, all these representations belong to the same node, thus they are inherently related to each other. We present two types of representation of nodes in the multi-graph, for each node, it has a \emph{shared representation}, which denotes the shared information in the multi-graph. Besides, it also has a \emph{specific representation} in each subgraph, which encodes the specific information in the subgraph.

\begin{figure*}
	\centering
	\includegraphics[height=86mm]{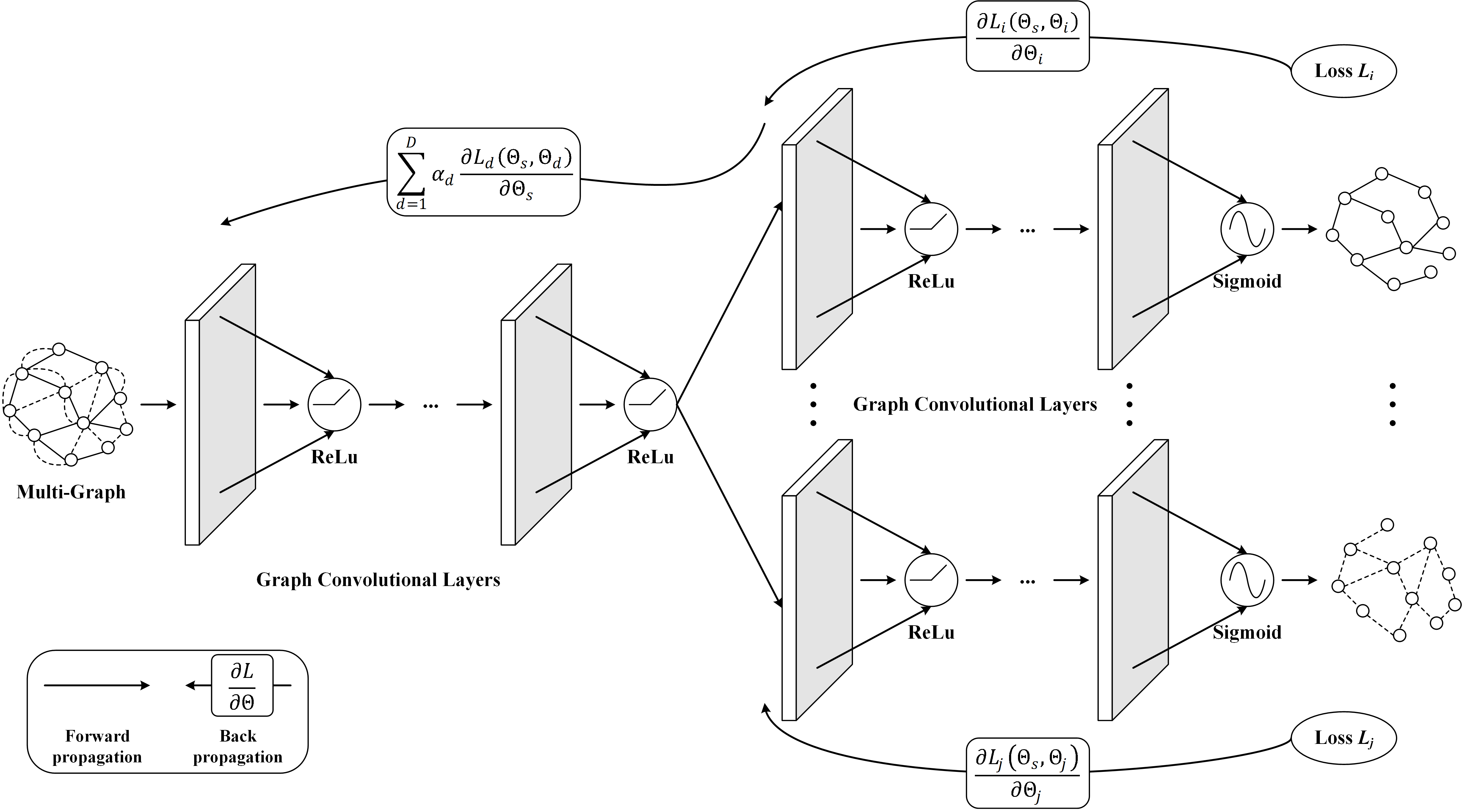}
	\caption{The architecture of DMGE.}
	\label{fig:DMGE}
\end{figure*}

To learn multiple types of node representations, the architecture of DMGE is presented in Figure~\ref{fig:DMGE}, which follows the multi-task learning regime~\cite{caruana1997multitask,ruder2017overview}. Specifically, the domain-shared layers are graph convolutional layers on multi-graph, which is used to learn \emph{shared representation} across domains. The domain-specific layers are also graph convolutional layers, which is used to learn \emph{specific representation} on each subgraph for each domain. The outputs of graph convolutional layers are node embeddings, which model the probability that an link existing between these nodes.

The graph convolutional layers can efficiently learn node embedding based on the neighborhood aggregation scheme. In DMGE, the shared graph convolutional layers are used to generate shared node embedding by encoding node attributes and multi-graph structure. Based on the shared embedding, the specific graph convolutional layers are used to generate specific node embedding on each subgraph by the same rule.


To learn shared embedding, the shared graph convolutional layers are defined as follows:

\begin{equation}
	\mathbf{X}_s^{(l+1)} = ReLu \left(\tilde{\mathbf{W}}^{-\frac{1}{2}} \tilde{\mathbf{A}} \tilde{\mathbf{W}}^{-\frac{1}{2}} \mathbf{X}_s^{(l)} \mathbf{\Theta}_s^{(l)}\right)
	\label{eq:shared}
\end{equation}
where $\mathbf{X}_s^{(l+1)} \in \mathbb{R}^{N \times E_s}$ is the shared embedding of multi-graph $\mathcal{G}$ in $l$-th layer, and $\mathbf{X}_s^{(0)} = \mathbf{X} \in \mathbb{R}^{N \times M}$ is the matrix of node feature. $\tilde{\mathbf{A}} = \mathbf{A} + \mathbf{I}_N \in \mathbb{R}^{N \times N}$ is the adjacency matrix of graph $\mathcal{G}$ with added self-connections, $\mathbf{A} \in \mathbb{R}^{N \times N}$ is the adjacency matrix, and $\mathbf{A}(i, j)=1$ if there are any links between node $v_i$ and $v_j$, and $\mathbf{I}_N$ is the identity matrix. $\tilde{\mathbf{W}} \in \mathbb{R}^{N \times N}$ is a diagonal matrix, and $\tilde{\mathbf{W}}(i, i)=\sum_{j} \tilde{\mathbf{A}}(i, j)$. $\mathbf{\Theta}_s^{(l)}$ is the shared weight matrix of $l$-th layer. The output of the $l$-th shared graph convolutional layer is the shared node embedding $\mathbf{X}_s$.

Based on the shared embedding, the specific graph convolutional layers are defined as follows:

\begin{equation}
\left\{
\begin{aligned}
&\mathbf{X}_1^{(l+2)} = ReLu \left(\tilde{\mathbf{W}}_1^{-\frac{1}{2}} \tilde{\mathbf{A}}_1 \tilde{\mathbf{W}}_1^{-\frac{1}{2}} \mathbf{X}_1^{(l+1)} \mathbf{\Theta}_1^{(l+1)}\right)\\
&\, \, \,  ...\\
&\mathbf{X}_D^{(l+2)} = ReLu \left(\tilde{\mathbf{W}}_D^{-\frac{1}{2}} \tilde{\mathbf{A}}_D \tilde{\mathbf{W}}_D^{-\frac{1}{2}} \mathbf{X}_D^{(l+1)} \mathbf{\Theta}_D^{(l+1)}\right)
\end{aligned}	
\right.
\label{eq:specific}
\end{equation}
where $\mathbf{X}_d^{(l+2)} \in \mathbb{R}^{N \times E}$ is the specific embedding of subgraph $\mathcal{G}_d$ in $l$$+$$1$-th layer, and $\mathbf{X}_d^{(l+1)} = \mathbf{X}_s \in \mathbb{R}^{N \times M}$ is the shared embedding. $\tilde{\mathbf{A}}_d = \mathbf{A}_d + \mathbf{I}_N \in \mathbb{R}^{N \times N}$ is the adjacency matrix of subgraph $\mathcal{G}_d$ with added self-connections, $\mathbf{A}_d \in \mathbb{R}^{N \times N}$ is the adjacency matrix, and $\mathbf{A}_d(i, j)$ is the weight of edge $(v_i, v_j)$. $\tilde{\mathbf{W}}_d \in \mathbb{R}^{N \times N}$ is a diagonal matrix, and $\tilde{\mathbf{W}}_d(i, i)=\sum_{j} \tilde{\mathbf{A}}_d(i, j)$. $\mathbf{\Theta}_d^{(l+1)}$ is the specific weight matrix of $l$$+$$1$-th layer. The output of the specific graph convolutional layers is the set of node embedding $\mathcal{X} = \{\mathbf{X}_1, ..., \mathbf{X}_D\}$.

To learn node embeddings, we train the neural network in an unsupervised manner by modeling the graph structure. We use the embeddings to generate the linkage between two nodes, i.e. the probability that there exists an edge between these nodes. Therefore, we formulate the embedding learning as a binary classification problem by using the embeddings of two nodes.

The probability that there exists an edge between node $v_i$ and node  $v_j$ in subgraph $\mathcal{G}_d$ is defined in Eq.~(\ref{eq:link}), and the probability that there exists no edge between node $v_i$ and node  $v_k$ in subgraph $\mathcal{G}_d$ is defined in Eq.~(\ref{eq:no_link}):

\begin{equation}
	p\left(1|v_i, v_j, d\right)=\sigma \left(\mathbf{x}_{d,i}^T \cdot \mathbf{x}_{d,j}\right)
	\label{eq:link}
\end{equation}

\begin{equation}
	\begin{aligned}
		p\left(0|v_i, v_k, d\right) = & 1-p\left(1|v_i, v_k, d\right)\\
		= & \sigma \left(-\mathbf{x}_{d,i}^T \cdot \mathbf{x}_{d,k}\right)
	\end{aligned}
	\label{eq:no_link}
\end{equation}
where $\mathbf{x}_{d, i}$ is the $i$-th row of $\mathbf{X}_d$, which is the embedding vector of node $v_i$ in subgraph $\mathcal{G}_d$. $\sigma(\cdot)$ is the sigmoid function.

Therefore, the objective is to generate the embedding by maximizing the log-likelihood function as follows:

\begin{equation}
	\begin{aligned}
		& \max \log \left(\prod_{v_j \in \mathcal{S}_{d, p}} p \left(1|v_i, v_j, d \right) \cdot \prod_{v_k \in \mathcal{S}_{d, n}} p \left(0|v_i, v_k, d \right)\right)\\
		= & \min - \! \sum_{v_j \in \mathcal{S}_{d, p}} \!\!\! \log p \left(1|v_i, v_j, d \right) - \! \sum_{v_k \in \mathcal{S}_{d, n}} \!\!\! \log p \left(0|v_i, v_k, d \right)\\
		= & \min - \! \sum_{v_j \in \mathcal{S}_{d, p}} \!\!\! \log \sigma \! \left(\mathbf{x}_{d,i}^T \cdot \mathbf{x}_{d,j}\right) - \! \sum_{v_k \in \mathcal{S}_{d, n}} \!\!\! \log \sigma \! \left(-\mathbf{x}_{d,i}^T \cdot \mathbf{x}_{d,k}\right)		
	\end{aligned}
	\label{eq:log_likelihood}
\end{equation}
where $\mathcal{S}_{d, p}$ is the set of positive samples in subgraph $\mathcal{G}_d$, which contains the tuples $(v_i, v_j, d)$ with an edge between node $v_i$ and node $v_j$ in subgraph $\mathcal{G}_d$. $\mathcal{S}_{d, n}=\{v_k|k=1, ..., S\}$ is the set of negative samples in subgraph $\mathcal{G}_d$. The negative samples are sampled from node set $\mathcal{V}$ by using negative sampling~\cite{mikolov2013efficient,mikolov2013distributed}, which contains the tuples $(v_i, v_k, d)$ with no edge between node $v_i$ and node $v_k$ in subgraph $\mathcal{G}_d$.

Therefore, the objective function is defined as Eq.~(\ref{eq:loss}), in which $L_d$ is the loss function of subgraph $\mathcal{G}_d$.

\begin{equation}
\!\!\! \left\{
\begin{aligned}
& L_1= \min - \!\!\!\! \sum_{v_j \in \mathcal{S}_{1, p}} \!\!\!\!\! \log \sigma \! \left(\mathbf{x}_{1,i}^T \cdot \mathbf{x}_{1,j}\right) - \!\!\!\!\! \sum_{v_k \in \mathcal{S}_{1, n}} \!\!\!\!\! \log \sigma \! \left(-\mathbf{x}_{1,i}^T \cdot \mathbf{x}_{1,k}\right) \\
& \, \, \,  ...\\
& L_D = \min - \!\!\!\!\!\! \sum_{v_j \in \mathcal{S}_{D, p}} \!\!\!\!\!\! \log \sigma \! \left(\mathbf{x}_{D,i}^T \cdot \mathbf{x}_{D,j}\right) - \!\!\!\!\!\!\! \sum_{v_k \in \mathcal{S}_{D, n}} \!\!\!\!\!\! \log \sigma \! \left(-\mathbf{x}_{D,i}^T \cdot \mathbf{x}_{D,k}\right) \\
\end{aligned}	
\right.
\label{eq:loss}
\end{equation}

\subsection{Optimization}
\label{sec:optimization}
In our model, the parameter set $\mathbf{\Theta}_s$ of shared graph convolutional layers is shared across domains, while parameter sets $\mathbf{\Theta}_d$ $(d=\{1, ..., D\})$ of specific graph convolutional layers are domain-specific. To train DMGE and benefit all domains, we need to optimize all the objectives $L_d(\mathbf{\Theta}_s, \mathbf{\Theta}_d)$. In multi-task learning~\cite{caruana1997multitask,ruder2017overview}, a commonly used method to optimize the objective function Eq.~(\ref{eq:loss}) is to solve the weighted summation of all $L_d$. However, stacking multiple layers brings additional difficulties to train the model~\cite{li2018deeper}, and it is time-consuming to tune the weight to obtain the optimal solution. Therefore, we formulate the problem as multi-objective optimization, and the optimization objective is defined as follows: 

\begin{equation}
\left\{
\begin{aligned}
&\min_{\mathbf{\Theta}_s, \mathbf{\Theta}_1}\, L_1(\mathbf{\Theta}_s, \mathbf{\Theta}_1)\\
&\, \, \,  ...\\
&\min_{\mathbf{\Theta}_s, \mathbf{\Theta}_D}\, L_D(\mathbf{\Theta}_s, \mathbf{\Theta}_D)
\end{aligned}	
\right.
\label{eq:multi-objective_optimization}
\end{equation}

The goal of Eq.~(\ref{eq:multi-objective_optimization}) is to find the solution which is optimal for each objective (i.e., each domain). To solve the multi-objective optimization, we introduce a multiple-gradient descent optimizer. Firstly, we state the Karush-Kuhn-Tucker (KKT) conditions~\cite{kuhn1951nonlinear} for the multi-objective optimization in Eq.~(\ref{eq:multi-objective_optimization}), which is a necessary condition for the optimal solution of multi-objective optimization:

\vspace{-0.1cm}
\begin{equation}
\left\{
\begin{aligned}
& \sum_{d=1}^{D} \alpha_d \frac{\partial L_d(\mathbf{\Theta}_s, \mathbf{\Theta}_d)}{\partial \mathbf{\Theta}_s} = 0\\
& \frac{\partial L_d(\mathbf{\Theta}_s, \mathbf{\Theta}_d)}{\partial \mathbf{\Theta}_d} = 0, (\forall d \in \{1, ..., D\})\\
& \sum_{d=1}^{D} \alpha_d = 1\\
& \alpha_d \geq 0, (\forall d \in \{1, ..., D\})			
\end{aligned}	
\right.
\label{eq:KKT}
\end{equation}
where $\alpha_d$ is the weight of objective $L_d(\mathbf{\Theta}_s, \mathbf{\Theta}_d)$.

As proved in~\cite{desideri2012multiple}, either the solution to Eq.~(\ref{eq:minimum-norm_point}) is 0 and the result satisfies the KKT conditions Eq.~(\ref{eq:KKT}), or the solution gives a descent direction that improves all objectives in Eq.~(\ref{eq:multi-objective_optimization}). Thus, solving the KKT conditions Eq.~(\ref{eq:KKT}) is equivalent to optimizing Eq.~(\ref{eq:minimum-norm_point}).

\begin{equation}
\begin{aligned}
\min_{\alpha_1, ..., \alpha_D} & \left\| \sum_{d=1}^{D} \alpha_d \frac{\partial L_d(\mathbf{\Theta}_s, \mathbf{\Theta}_d)}{\partial \mathbf{\Theta}_s} \right\|_2^2\\
s.t. & \left\{
\begin{aligned}
& \sum_{d=1}^{D} \alpha_d = 1\\
& \alpha_d \geq 0, (\forall d \in \{1, ..., M\})
\end{aligned}
\right.			
\end{aligned}
\label{eq:minimum-norm_point}	
\end{equation}


To clearly illustrate how the optimizer works, we consider the case of two domains. The optimization objective Eq.~(\ref{eq:minimum-norm_point}) can be simplified as:

\begin{equation}
\begin{aligned}
\min_{\alpha} & \, \left\| \alpha \frac{\partial L_1(\mathbf{\Theta}_s, \mathbf{\Theta}_1)}{\partial \mathbf{\Theta}_s} + (1-\alpha) \frac{\partial L_2(\mathbf{\Theta}_s, \mathbf{\Theta}_2)}{\partial \mathbf{\Theta}_s} \right\|_2^2\\
s.t. & \,\, 0 \leq \alpha \leq 1
\end{aligned}
\label{eq:two_domains}	
\end{equation}
where $\alpha$ is the weight of $L_1(\Theta_s, \Theta_1)$.


The Eq.~(\ref{eq:two_domains}) is a unary quadratic equation of $\alpha$, and the solution to Eq.~(\ref{eq:two_domains}) is:

\begin{equation}
\alpha =
\begin{cases}
0, & \boldsymbol{\rm{U}}^T \boldsymbol{\rm{V}} \geq \boldsymbol{\rm{V}}^T \boldsymbol{\rm{V}}\\
1, & \boldsymbol{\rm{U}}^T \boldsymbol{\rm{V}} \geq \boldsymbol{\rm{U}}^T \boldsymbol{\rm{U}}\\
\frac{\left(\boldsymbol{\rm{V}} - \boldsymbol{\rm{U}}\right)^T \boldsymbol{\rm{V}}}{\left\|\boldsymbol{\rm{U}} - \boldsymbol{\rm{V}} \right\|_2^2}, & else
\end{cases}
\label{eq:alpha}
\end{equation}
where $\boldsymbol{\rm{U}} = \frac{\partial L_1\left(\mathbf{\Theta}_s, \mathbf{\Theta}_1\right)}{\partial \mathbf{\Theta}_s}$, $\boldsymbol{\rm{V}} = \frac{\partial L_2\left(\mathbf{\Theta}_s, \mathbf{\Theta}_2\right)}{\partial \mathbf{\Theta}_s}$.

With the weight $\alpha$, we update the parameters of the model as follows, as the parameter sets $\mathbf{\Theta}_1$ and $\mathbf{\Theta}_2$ are domain-specific, we update $\mathbf{\Theta}_1$ and $\mathbf{\Theta}_2$ by using Eq.~(\ref{eq:domain-specific}) for each domain, respectively. The parameter set $\mathbf{\Theta}_s$ is shared across domains, thus we apply the weighted gradient $\alpha \frac{\partial L_1(\mathbf{\Theta}_s, \mathbf{\Theta}_1)}{\partial \mathbf{\Theta}_s} + (1 - \alpha) \frac{\partial L_2(\mathbf{\Theta}_s, \mathbf{\Theta}_2)}{\partial \mathbf{\Theta}_s}$ as a gradient update to the shared parameters as defined in Eq.~(\ref{eq:domain-shared}). Notice that $\alpha$ is trained by the optimizer according to the gradient of each domain.\\


\begin{equation}
\mathbf{\Theta}_d = \mathbf{\Theta}_d - \eta \frac{\partial L_d(\mathbf{\Theta}_s, \mathbf{\Theta}_d)}{\partial \mathbf{\Theta}_d}
\label{eq:domain-specific}
\end{equation}

\begin{equation}
\mathbf{\Theta}_s = \mathbf{\Theta}_s - \eta \sum_{d=1}^{D} \alpha_d \frac{\partial L_d(\mathbf{\Theta}_s, \mathbf{\Theta}_d)}{\partial \mathbf{\Theta}_s}
\label{eq:domain-shared}
\end{equation}

Finally, we summarize the learning procedure of DMGE in Algorithm~\ref{alg:algorithm}. In DMGE, the input includes the multi-graph $\mathcal{G}$ and the node feature matrix $\mathbf{X}$. In line 1, we first initialize the parameter sets $\mathbf{\Theta}_s$, $\mathbf{\Theta}_d$ and the weight $\alpha_d$ of each domain. Then, we operate convolution on the multi-graph $\mathcal{G}$ in line 2. For each subgraph $\mathcal{G}_d$, we sample a set of negative samples in line 5. We use the link information to train DMGE, compute the gradients and update the parameters of specific graph convolutional layers in line 6, and we compute the gradients of shared graph convolutional layers in line 7. Based on the gradients, we compute $\alpha_d$ in line 8, and update the parameters of shared graph convolutional layers in line 9. Finally, we return a set of node embeddings in line 12. 

\begin{algorithm}[tb]
	\caption{Deep Multi-Graph Embedding}
	\label{alg:algorithm}
	
	\begin{flushleft}
		\textbf{Input}:
		\setlength{\hangindent}{3em}
		A multi-graph $\mathcal{G}=(\mathcal{V}, \left\{\mathcal{E}_1, ..., \mathcal{E}_D\right\})$, and the node feature matrix $\mathbf{X} \in \mathbb{R}^{N \times M}$ \\
		\textbf{Parameter}: 
		\setlength{\hangindent}{3em}
		$\mathbf{\Theta}_s$, $\mathbf{\Theta}_d$ $(d=\{1, ..., D\})$, and $\alpha_d$\\
		\textbf{Output}: 
		A set of node embedding $\mathcal{X} = \{\mathbf{X}_1, ..., \mathbf{X}_D\}$ $(\mathbf{X}_d \in \mathbb{R}^{N \times E})$
	\end{flushleft}

	\begin{algorithmic}[1] 
		\STATE Initialize parameters $\mathbf{\Theta}_s$, $\mathbf{\Theta}_d$, and $\alpha_d$.
		\STATE Operate convolution on multi-graph $\mathcal{G}$ by Eq.~(\ref{eq:shared}) and Eq.~(\ref{eq:specific}).
		\FOR{$d \in [1, D]$}
		\FOR{$(v_i, v_j) \in \mathcal{E}_d$}
		\STATE Sample a set of negative samples $\mathcal{S}_{d, n}$.
		\STATE Update $\mathbf{\Theta}_d = \mathbf{\Theta}_d - \eta \frac{\partial L_d(\mathbf{\Theta}_s, \mathbf{\Theta}_d)}{\partial \mathbf{\Theta}_d}$.
		\STATE Compute gradients of $\mathbf{\Theta}_s$: $\frac{\partial L_d(\mathbf{\Theta}_s, \mathbf{\Theta}_d)}{\partial \mathbf{\Theta}_s}$.
		\STATE Compute $\alpha_d$ by using Eq.~(\ref{eq:alpha}).
		\STATE Update $\mathbf{\Theta}_s = \mathbf{\Theta}_s - \eta \sum_{d=1}^{D} \alpha_d \frac{\partial L_d(\mathbf{\Theta}_s, \mathbf{\Theta}_d)}{\partial \mathbf{\Theta}_s}$.
		\ENDFOR
		\ENDFOR
		\STATE \textbf{return} A set of node embedding $\mathcal{X} = \{\mathbf{X}_1, ..., \mathbf{X}_D\}$
	\end{algorithmic}
\end{algorithm}

\section{Experiments}

In this section, we first present the research questions about DMGE. Then, we introduce the datasets and experimental settings. Finally, we present the experimental results to demonstrate the effectiveness of DMGE.

We first present the following three research questions:

\begin{itemize}	
	\item \textbf{RQ1:} How does DMGE perform in the recommendation task compared with other state-of-the-art embedding methods for recommendation?
	
	\item \textbf{RQ2:} How does the parameter sensitivity affect the performance of DMGE for recommendation?
	
	\item \textbf{RQ3:} How does DMGE perform in the classic task on graph (e.g., link prediction) compared with other state-of-the-art graph embedding methods?
\end{itemize}

\subsection{Datasets}
We evaluate our model on two real-world datasets, the details of datasets are as follows and the statistics of datasets are presented in Table~\ref{tab:dataset}.

\begin{table}
	\caption{Statistics of datasets}
	\label{tab:dataset}
	\begin{tabular}{c|c|c|c}
		\toprule
		Dataset & Domain/Relation & Nodes & Edges\\
		\midrule
		\multirow{2}*{Tencent App Store} & Homepage & 18,229 & 548,930 \\
		~ & Search & 18,229 & 936,065 \\	
		\hline		
		\multirow{2}*{Youtube} & Friendship &  15,088 &  76,765 \\
		~ & Co-friends &  15,088 &  1,940,806 \\	
		\bottomrule
	\end{tabular}
\end{table}

\begin{itemize}	
	\item \textbf{Tencent App Store}: It is the App download records from a company App store, which contains recommendation domain and search domain. The time span of the dataset is 31 days, the number of Apps is 18,229, and the number of user is 1,011,567. Based on users' download records, we construct the item graph for each domain, and the statistics of graph is presented in Table~\ref{tab:dataset}. We use this dataset for App recommendation task. 
	
	\item \textbf{Youtube}\footnote{\url{http://socialcomputing.asu.edu/datasets/YouTube}}:  YouTube dataset~\cite{Zafarani+Liu:2009} consists of two types of relation among users, i.e. friendship and co-friends. Specifically, the friendship relation means two users are friends, and the co-friends means two users have shared friends. We use this dataset for link prediction task.
\end{itemize}

\subsection{Experimental Settings}

\subsubsection{Baseline Methods}
For both tasks, we choose the following state-of-the-art graph embedding methods as baselines:

\begin{itemize}
	\item \textbf{DeepWalk}~\cite{perozzi2014deepwalk}: It applies random walk on graph to generate node sequences, and uses Skip-Gram algorithm to learn embedding. We apply DeepWalk to each subgraph separately.
	
	\item \textbf{LINE}~\cite{tang2015line}: It learns node embedding through preserving both local and global graph structures. We apply LINE to each subgraph separately. 
	
	\item \textbf{node2vec}~\cite{grover2016node2vec}: It designs a biased random walk procedure, and can explore diverse neighborhoods. We apply node2vec to each subgraph separately.
	
	\item \textbf{GCN}~\cite{kipf2017semi}: It operates convolution on graph, and can generate node embedding based on neighborhoods. We apply GCN to each subgraph separately.
	
	\item \textbf{mGCN}~\cite{ma2019multi}: It applies graph convolutional networks for multi-graph embedding. It can generate both general embeddings to capture the information for nodes over the entire graph and dimension-specific embeddings to capture the information for nodes in each subgraph.
	
	\item \textbf{DMGE ($\alpha$)}: It is a variant of DMGE. It defines the objective function as the weighted summation of $L_d$ in Eq.~(\ref{eq:multi-objective_optimization}) for multi-graph embedding, in which $\alpha$ is the weight of the first domain. 
\end{itemize}

For the App recommendation task, besides the above baselines, we also compare with the matrix factorization (\textbf{MF})~\cite{koren2009matrix}, which factorizes user-item matrix into user embedding and item embedding, respectively. We apply MF to each domain separately.


\subsubsection{Evaluation Metrics}
To evaluate the performance of recommendation, we compare the recommended top-$N$ list $R_u$ with the corresponding ground truth list $T_u$ for each user $u \in \mathcal{U}$, and use the following metrics to evaluate the top-$N$ recommended results:

\begin{itemize}	
	\item \textbf{Recall@$N$}: It calculates the fraction of the ground truth (i.e., the user downloaded Apps) that are recommended by different algorithms in Eq.~(\ref{eq:recall}), where $\mathcal{U}$ is the user set, $h_u$ denotes the number of downloaded Apps hits in the candidate top-$N$ App list $R_u$ for user $u$, and $t_u$ denotes the number of downloaded App list $T_u$ of user $u$. A larger value of recall@$N$ means better performance.
	
	\begin{equation}
		Recall@N = \frac{\sum_{u \in \mathcal{U}} h_{u}}{\sum_{u \in \mathcal{U}} t_{u}}
		\label{eq:recall}
	\end{equation}
	
	\item \textbf{MRR@$N$}: Mean Reciprocal Rank (MRR) uses the multiplicative inverse of the rank of the first hit item among top-$N$ item list to evaluate the performance of rank in Eq.~(\ref{eq:MRR}), where $r_u$ is the rank of the first hit item. A larger value of MRR@$N$ means better performance.
	
	\begin{equation}
		MRR@N = \frac{1}{|\mathcal{U}|} \sum_{u \in \mathcal{U}} \frac{1}{r_u}
		\label{eq:MRR}
	\end{equation}

\end{itemize}

To evaluate the performance of link prediction, we use the metrics of binary classification: AUC and F1.

\subsubsection{Model Parameters}

The parameters of DMGE are set as follows: 

\begin{itemize}
	\item \emph{Network architecture}. The number of shared and specific graph convolutional layers are both 1, shared hidden size is 64, and specific hidden size is 16.
	
	\item \emph{Initialization}. The node feature matrix can be initialized randomly, or by other embedding methods, we initialize it as the identity matrix.
	
	\item \emph{Gradient normalization}. We normalize the gradient of shared parameter $\mathbf{\Theta}_s$ of each domain, and then use the normalized gradient to calculating $\alpha$ in Eq.~(\ref{eq:alpha}). The normalized gradient of domain $d$ is $\mathbf{G}_d / \left(\left\| \mathbf{G}_d \right\|_2 \cdot L_d \right)$, where $\mathbf{G}_d =  \frac{\partial L_d\left(\mathbf{\Theta}_s, \mathbf{\Theta}_d\right)}{\partial \mathbf{\Theta}_s}$ is the unnormalized gradient.
	
	
	\item \emph{Other hyper-parameters}. The number of negative samples is 2; the embedding dimension is 16; the dropout of shared graph convolutional layers is 0.3 and that is 0.1 of specific graph convolutional layers; the batch size is 256 and we train the model for a maximum of 10 epochs using Adam.
\end{itemize}

In all methods, the dimension of embedding is set to 16. The parameters of baselines are fine-tuning, and set as follows:

\begin{itemize}	
	\item MF. It is implemented using LibMF\footnote{https://www.csie.ntu.edu.tw/~cjlin/libmf/}. 
	
	\item DeepWalk. The length of context window is 5; the length of random walk is 20; the number of walks per node is 50.
	
	\item LINE. The number of negative samples is 2.
	
	\item node2vec.  The length of context window is 5; the length of random walk is 20; the number of walks per node is 50; the number of negative samples is 2; $p$ is 1 and $q$ is 0.25.
	
	\item GCN. The number of graph convolutional layers is 1.
	
	\item mGCN. The initial general representation size is 64, other parameter settings are the same as~\cite{ma2019multi}, and we train the model for a maximum of 20 epochs using Adam.
	
	\item DMGE ($\alpha$). Considering that both domains are important, we set the weight $\alpha$ to 0.5; the other parameter settings are the same as DMGE.
\end{itemize}


\subsection{Embedding for Recommendation}
To demonstrate the performance of DMGE in recommendation task (\textbf{RQ1}), we compare DMGE with other state-of-the-art embedding methods. The intuition is that learning better item embeddings will achieve better performance of recommendation.

Generally, users' preferences can be characterized by the items they have interacted with, thus 
we represent users by aggregating embeddings of their interacted items. There are several ways to aggregate item embeddings, such as: average~\cite{zhao2018learning}, RNN~\cite{okura2017embedding}. We apply average here, and represent users by using the average item embeddings of their interacted items:

\begin{equation}
\mathbf{u}_{d} = \frac{1}{I_d} \sum_{i=1}^{I_d} \mathbf{x}_{d, i}
\end{equation}
where $\mathbf{u}_d$ is the embedding of user $u \in \mathcal{U}$ in domain $d$, $I_d$ is the number of items user $u$ has interacted with, and $\mathbf{x}_{d, i}$ is the embedding of item $i$ in domain $d$.

For each domain, we measure user-item similarity by computing the cosine distance between user embedding and item embedding. Based on the user-item similarity, we then generate candidate top-$N$ items for each user. We use consecutive 26 days data to train item embedding, and measure the performance of recommendation in the next 5 days by using the metric Recall@$N$ and MRR@$N$. The performance of different methods for recommendation domain and search domain is presented in Table~\ref{tab:recommendation_recall}, Table~\ref{tab:search_recall}, Table~\ref{tab:recommendation_mrr} and Table~\ref{tab:search_mrr}. (Note that the best results are indicated by the bold font.)

Based on the results, we have the following observations:

\begin{itemize}
	\item We first compare the performance of single-domain methods, including: MF, DeepWalk, LINE, node2vec and GCN. We can observe the graph embedding methods outperforms MF, as MF only takes into account the explicit user-item interactions, while ignoring item co-occurrences in users' behaviors, which can be captured by graph embedding methods.
	
	\item The overall performance of cross-domain methods (i.e., mGCN, DMGE ($\alpha$), DMGE) is better than the single domain methods, which demonstrates that fusing information from correlated domains is helpful to learn better cross-domain representation, and can improve the performance of recommendation in both domains. When $N$ is less than 40 in recommendation domain and $N$ is less than 30 in search domain, the Recall of mGCN is worse than the single domain methods, the possible reason is that weight between within-domain and across-domain in mGCN is a hyper-parameter to be tuned, and can not be adaptively learned by the importance of each domain. Both DMGE and DMGE ($\alpha$) are consistently outperforms the single domain methods.
	
	\item Compared the cross-domain embedding methods, both DMGE ($\alpha$) and DMGE outperform mGCN, which indicates that our proposed graph neural network is effective to learn better representations.
	
	\item DMGE outperforms DMGE ($\alpha$) (except Recall@1000 in recommendation domain). The average of $\alpha$ in DMGE is 0.4409, thus when $\alpha=0.5$, DMGE ($\alpha$) can also achieve good performance. However, in DMGE ($\alpha$), it is time-consuming and computationally expensive to tune the hyper-parameter $\alpha$ to obtain the optimal result. While in DMGE, $\alpha$ is a trainable parameter. Thus, we recommend to use the multiple-gradient descent optimizer to train the model.
\end{itemize}

Overall, the proposed DMGE outperforms the state-of-the-art embedding methods, and improves the performance of recommendation in both domains.

\begin{table*}
	\caption{Recall@$N$ Performance of Different Methods in Recommendation Domain}
	\label{tab:recommendation_recall}

	\begin{tabular}{c|c|ccccccccccc}
		\toprule
		Domain & Recall@$N$ & 10 & 20 & 30 & 40 &50 & 60 & 70& 80 & 90 & 100 & 1000 \\
		\midrule
		\midrule
		\multirow{5}*{Single} & MF & 0.0301 & 0.0453 & 0.0565 & 0.0658 & 0.0739 & 0.0812 & 0.0880 & 0.0942 & 0.1002 & 0.1065 & 0.2932 \\

		~& DeepWalk & 0.0730 & 0.1104 & 0.1363 & 0.1558 & 0.1720 & 0.1853 & 0.1975 & 0.2082 & 0.2186 & 0.2273 & 0.4744 \\
		
		~& LINE & 0.0471 & 0.0728 & 0.0933 & 0.1106 & 0.1258 & 0.1395 & 0.1525 & 0.1642 & 0.1754 & 0.1861 & 0.4977 \\
		
		~& node2vec & 0.0345& 0.0579 & 0.0773 & 0.0936 & 0.1080 & 0.1207 & 0.1324 & 0.1436 & 0.1534 & 0.1630 & 0.4574 \\

		~& GCN & 0.0743 & 0.1078 & 0.1317 & 0.1516 & 0.1688 & 0.1848 & 0.1977 & 0.2098 & 0.2217 & 0.2321 & 0.5624 \\
		\midrule
		\midrule
		\multirow{3}*{Cross} & mGCN & 0.0431 & 0.0835 & 0.1273 & 0.1677 & 0.2002 & 0.2142 & 0.2261 & 0.2383 & 0.2505 & 0.2627 &  0.6323\\

		~& DMGE ($\alpha=0.5$) & 0.1019 & 0.1607 & 0.2069 & 0.2436 & 0.2762 & 0.3035 & 0.3260 & 0.3471 & 0.3660 & 0.3826 & \textbf{0.7016} \\
	
		~& DMGE & \textbf{0.1024} & \textbf{0.1661} & \textbf{0.2109} & \textbf{0.2455} & \textbf{0.2767} & \textbf{0.3042} & \textbf{0.3277} & \textbf{0.3484} & \textbf{0.3669} & \textbf{0.3831} & 0.6929 \\
		\bottomrule
	\end{tabular}
\end{table*}



\begin{table*}
	\caption{Recall@$N$ Performance of Different Methods in Search Domain}
	\label{tab:search_recall}
	
	\begin{tabular}{c|c|ccccccccccc}
		\toprule
		Domain & Recall@$N$ & 10 & 20 & 30 & 40 &50 & 60 & 70& 80 & 90 & 100 & 1000 \\
		\midrule
		\midrule
		\multirow{5}*{Single} & MF & 0.0150 & 0.0251 & 0.0335 & 0.0408 & 0.0474 & 0.0533 & 0.0589 & 0.0642 & 0.0691 & 0.0739 & 0.2679 \\
		
		~& DeepWalk & 0.0638 & 0.1043 & 0.1338 & 0.1571 & 0.1761 & 0.1924 & 0.2064 & 0.2185 & 0.2291 & 0.2387 & 0.4676 \\
		
		~& LINE & 0.0392 & 0.0603 & 0.0769 & 0.0909 & 0.1030 & 0.1138 & 0.1238 & 0.1331 & 0.1414 & 0.1495 & 0.4293 \\
		
		~& node2vec & 0.0289 & 0.0471 & 0.0622 & 0.0753 & 0.0870 & 0.0982 & 0.1082 & 0.1174 & 0.1260 & 0.1346 & 0.4176 \\
		
		~& GCN & 0.0499 & 0.0764 & 0.0956 & 0.1111 & 0.1242 & 0.1363 & 0.1472 & 0.1575 & 0.1666 & 0.1754 & 0.4905 \\
		\midrule
		\midrule
		\multirow{3}*{Cross} & mGCN & 0.0478 & 0.0939 & 0.1454 & 0.1938 & 0.2218 & 0.2328 & 0.2399 & 0.2480 & 0.2565 & 0.2653 &  0.5920\\
		
		~& DMGE ($\alpha=0.5$) & 0.0823 & 0.1363 & 0.1784 & 0.2134 & 0.2415 & 0.2652 & 0.2857 & 0.3037 & 0.3206 & 0.3360 & 0.6254\\

		~& DMGE & \textbf{0.0885} & \textbf{0.1467} & \textbf{0.1900} & \textbf{0.2238} & \textbf{0.2517} & \textbf{0.2759} & \textbf{0.2971} & \textbf{0.3162} & \textbf{0.3328} & \textbf{0.3473} & \textbf{0.6263} \\
		\bottomrule
	\end{tabular}
\end{table*}



\begin{table*}
	\caption{MRR@$N$ Performance of Different Methods in Recommendation Domain}
	\label{tab:recommendation_mrr}
	
	\begin{tabular}{c|c|ccccccccccc}
		\toprule
		Domain & Recall@$N$ & 10 & 20 & 30 & 40 &50 & 60 & 70& 80 & 90 & 100 & 1000 \\
		\midrule
		\midrule
		\multirow{5}*{Single} & MF & 0.0149 & 0.0170 & 0.0180 & 0.0185 & 0.0188 & 0.0191 & 0.0193 & 0.0194 & 0.0196 & 0.0197 & 0.0208 \\
		
		~& DeepWalk & 0.0510 & 0.0549 & 0.0563 & 0.0571 & 0.0575 & 0.0578 & 0.0581 & 0.0582 & 0.0584 & 0.0585 & 0.0594 \\
		
		~& LINE & 0.0371 & 0.0398 & 0.0410 & 0.0417 & 0.0421 & 0.0424 & 0.0427 & 0.0429 & 0.0431 & 0.0432 & 0.0444 \\
			
		~& node2vec & 0.0265 & 0.0290 & 0.0302 & 0.0308 & 0.0312 & 0.0315 & 0.0317 & 0.0319 & 0.0321 & 0.0322 & 0.0334 \\
		
		~& GCN & 0.0558 & 0.0592 & 0.0606 & 0.0613 & 0.0619 & 0.0622 & 0.0625 & 0.0627 & 0.0628 & 0.0630 & 0.0642 \\
		\midrule
		\midrule
		\multirow{3}*{Cross} & mGCN & 0.0264 & 0.0311 & 0.0338 & 0.0354 & 0.0364 & 0.0367 & 0.0370 & 0.0372 & 0.0373 & 0.0375 &  0.0389\\
		
		~& DMGE ($\alpha=0.5$) & 0.0697 & 0.0756 & 0.0780 & 0.0793 & 0.0801 & 0.0807 & 0.0811 & 0.0814 & 0.0816 & 0.0817 & 0.0829 \\

		~& DMGE & \textbf{0.0699} & \textbf{0.0761} & \textbf{0.0785} & \textbf{0.0797} & \textbf{0.0805} & \textbf{0.0810} & \textbf{0.0814} & \textbf{0.0817} & \textbf{0.0819} & \textbf{0.0821} & \textbf{0.0832} \\
		\bottomrule
	\end{tabular}
\end{table*}



\begin{table*}
	\caption{MRR@$N$ Performance of Different Methods in Search Domain}
	\label{tab:search_mrr}
	
	\begin{tabular}{c|c|ccccccccccc}
		\toprule
		Domain & Recall@$N$ & 10 & 20 & 30 & 40 &50 & 60 & 70& 80 & 90 & 100 & 1000 \\
		\midrule
		\midrule
		\multirow{5}*{Single} & MF & 0.0120 & 0.0134 & 0.0141 & 0.0145 & 0.0148 & 0.0150 & 0.0151 & 0.0152 & 0.0153 & 0.0154 & 0.0164 \\		
		
		~& DeepWalk & 0.0468 & 0.0515 & 0.0534 & 0.0543 & 0.0549 & 0.0553 & 0.0556 & 0.0558 & 0.0560 & 0.0561 & 0.0571 \\
		
		~& LINE & 0.0345 & 0.0372 & 0.0384 & 0.0390 & 0.0395 & 0.0398 & 0.0400 & 0.0402 & 0.0403 & 0.0405 & 0.0417 \\
		
		~& node2vec & 0.0237 & 0.0260 & 0.0271 & 0.0278 & 0.0283 & 0.0286 & 0.0289 & 0.0290 & 0.0292 & 0.0293 & 0.0307 \\
		
		~& GCN & 0.0405 & 0.0437 & 0.0450 & 0.0457 & 0.0462 & 0.0465 & 0.0468 & 0.0470 & 0.0471 & 0.0473 & 0.0486 \\
		\midrule
		\midrule
		\multirow{3}*{Cross} & mGCN & 0.0324 & 0.0382 & 0.0415 & 0.0435 & 0.0443 & 0.0446 & 0.0448 & 0.0449 & 0.0450 & 0.0452 &  0.0465\\
		
		~& DMGE ($\alpha=0.5$) & 0.0592 & 0.0653 & 0.0678 & 0.0691 & 0.0700 & 0.0705 & 0.0709 & 0.0712 & 0.0714 & 0.0716 & 0.0728 \\

		~& DMGE & \textbf{0.0629} & \textbf{0.0693} & \textbf{0.0718} & \textbf{0.0731} & \textbf{0.0739} & \textbf{0.0744} & \textbf{0.0748} & \textbf{0.0751} & \textbf{0.0753} & \textbf{0.0755} & \textbf{0.0766} \\
		\bottomrule
	\end{tabular}
\end{table*}



\subsection{Parameter Sensitivity}
\begin{figure}
	\centering
	\includegraphics[width=85mm]{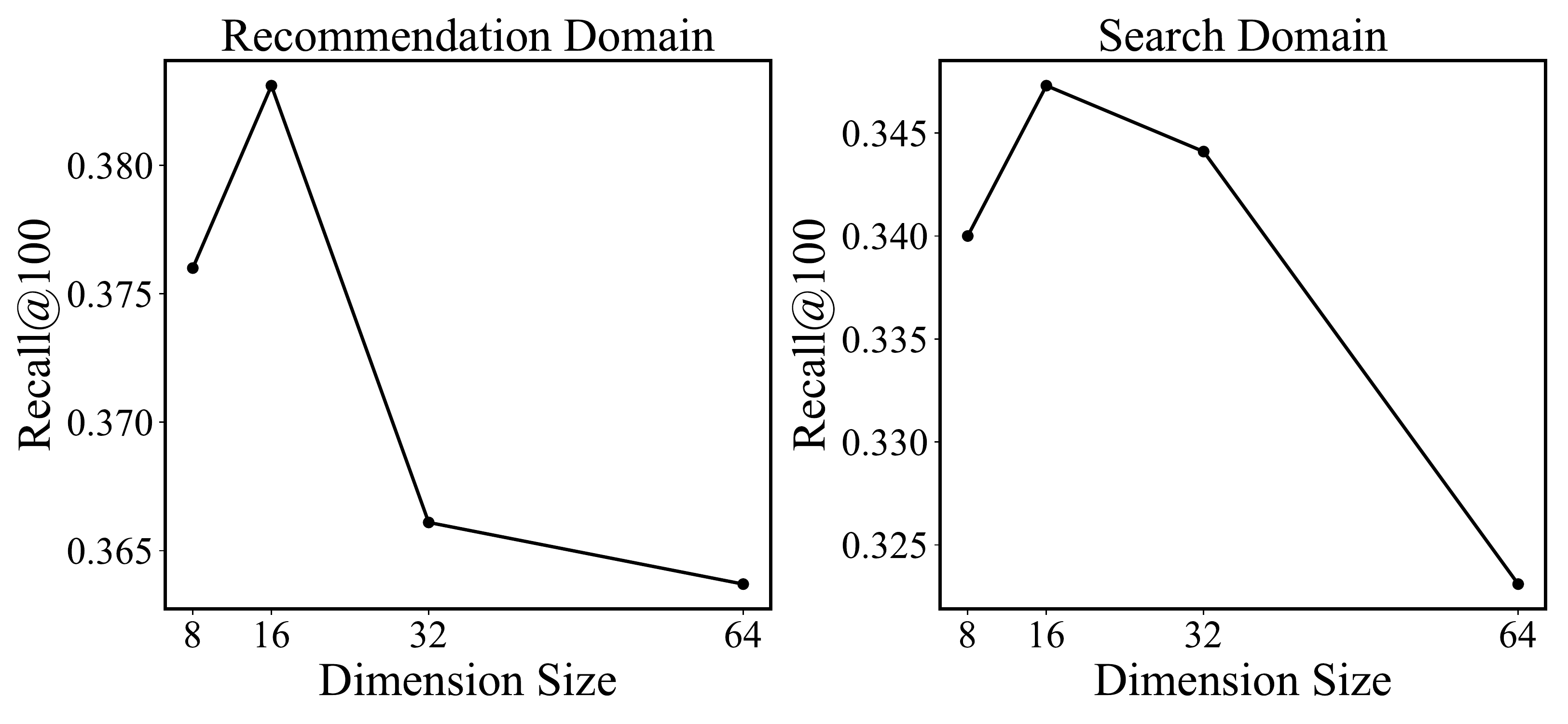}
	\caption{The parameter sensitivity analysis of embedding dimension for recommendation and search domain.}
	\label{fig:embedding_dimension}
\end{figure}

The key parameter that affects the performance of embedding is the dimension size (\textbf{RQ2}), we analyze how does the dimension size of learned embedding in DMGE affect the performance of recommendation. In particular, we test the dimension size $= \{8, 16, 32, 64\}$. Figure~\ref{fig:embedding_dimension} show the results of different embedding dimension in recommendation and search domain, and the evaluation metric is Recall@$100$.

As shown in Figure~\ref{fig:embedding_dimension}, in both domains, when the dimension of embedding is 16, our model performs best regarding the metric Recall@100. Therefore, we set the dimension of embedding as 16.

\subsection{Link Prediction}
To demonstrate the performance of DMGE in the link prediction task (\textbf{RQ3}), we compare DMGE with other state-of-the-art graph embedding methods. The intuition is that learning better node embeddings will achieve better performance of link prediction.

In the multi-graph, we perform link prediction in different subgraph separately. In each subgraph, we randomly remove 30\% of edges, and we aim to predict whether these removed edges exist. We formulate the link prediction task as a binary classification problem by using the embeddings of two nodes, and there are two types of combination: element-wise addition, element-wise multiplication.

\begin{figure}
	\centering
	\subfigure[AUC of different methods.]{
		\begin{minipage}{0.5\textwidth}
			\centering
			\includegraphics[width=85mm]{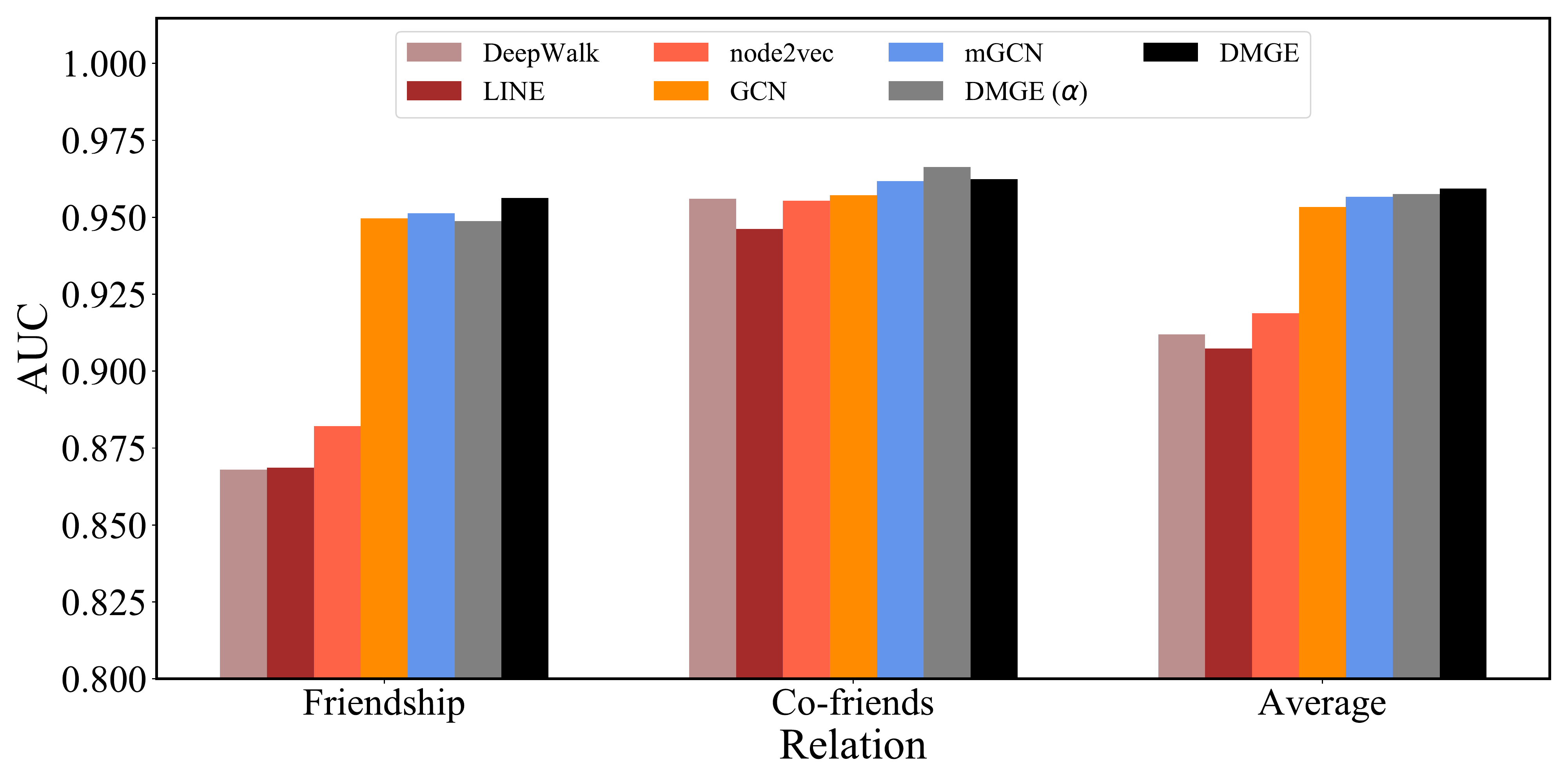}
	\end{minipage}}
	\subfigure[F1 of different methods.]{
		\begin{minipage}{0.5\textwidth}
			\centering
			\includegraphics[width=85mm]{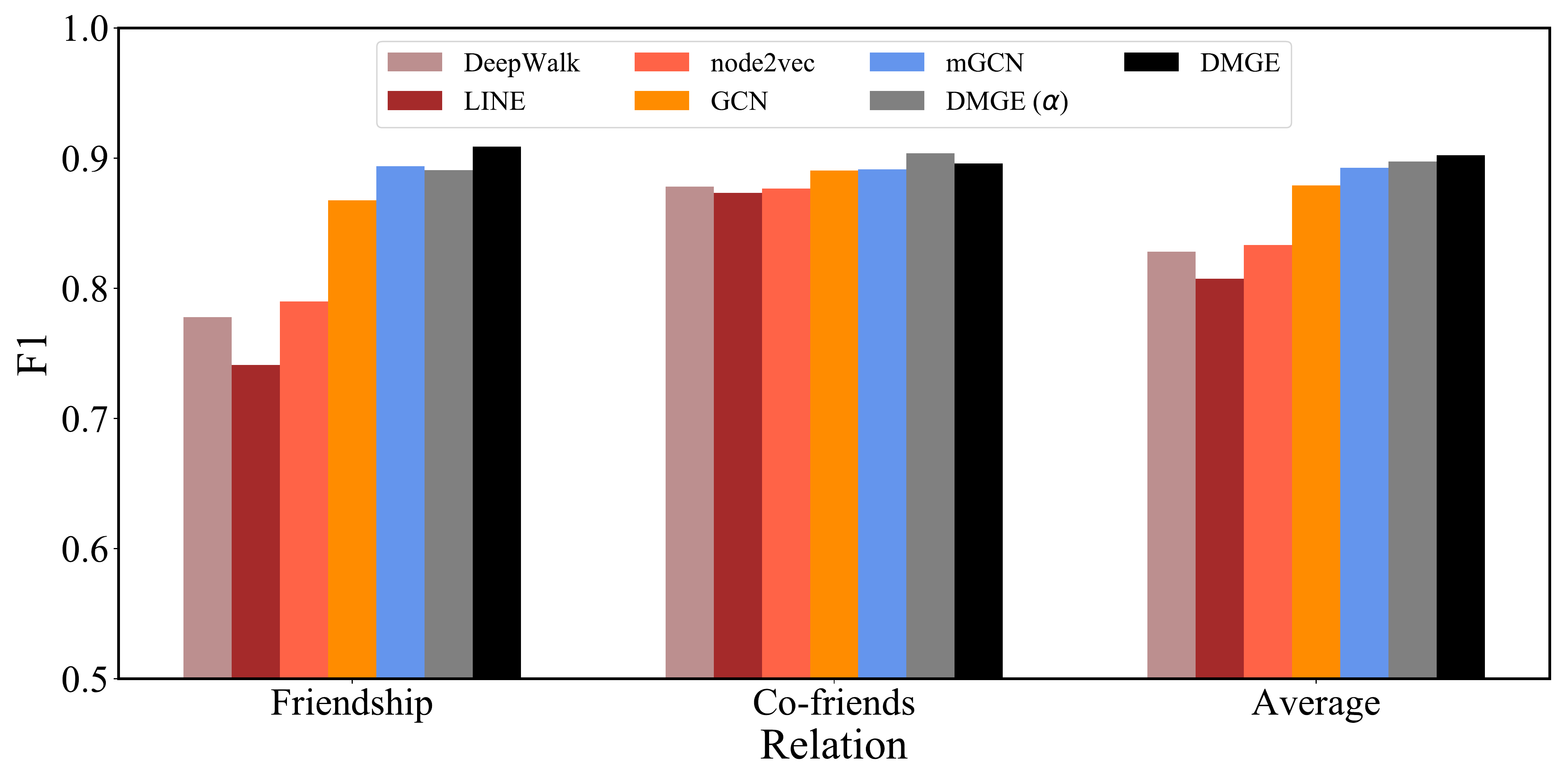}
	\end{minipage}}
	\caption{The performance of different methods in link prediction.}
	\label{fig:link_prediction}
\end{figure}

In training set, we use the remaining node pairs as positive samples, and randomly sample an equal number of not connected node pairs as negative samples. In testing set, we use the removed node pairs as positive samples, and randomly sample an equal number of not connected node pairs as negative samples. We train a binary classifier using logistic regression on the training set, and evaluate the performance of link prediction on the testing set. For each method, we select the optimal combination of embeddings and present the best results. The results of different methods are presented in Figure~\ref{fig:link_prediction}, including the results of each relation, and the average performance over all dimensions. 

Based on the results, we have the following observations:

\begin{itemize}
	\item The multi-graph embedding methods (i.e., mGCN, DMGE ($\alpha$), DMGE) outperforms the single graph embedding methods (i.e., DeepWalk, LINE, node2vec, GCN), which indicates that using multiple relations in the multi-graph is helpful to learn better representation.
	
	\item DMGE ($\alpha$) and DMGE outperform mGCN, which indicates that our proposed graph neural network is effective to learn better representations.
	
	\item The average performance of DMGE is better than DMGE ($\alpha$), which indicates the effectiveness of training the model using multi-objective optimization. 
\end{itemize}

\subsection{Discussion}
\label{sec:analysis}

\subsubsection{The Usage of Embedding} 
The embeddings of DMGE can be used for candidate items generation in the recall stage. Through calculating the pairwise similarities between the embeddings of users and items, we can generate a candidate set of items which users may like, and the candidate set can be further used in the ranking stage to generate the final recommendation set of items~\cite{covington2016deep}. Besides, the embeddings can also be used for transfer learning~\cite{ni2018perceive}, and alleviating the sparsity and cold start problem~\cite{zhao2018learning}.


\subsubsection{Cross-Domain Representation Leaning}
Not only designed for two domains, DMGE can also easily be extended to more domains. Using the Frank-Wolfe algorithm~\cite{frank1956algorithm,jaggi2013revisiting}, we can solve the optimization problem in Eq.~(\ref{eq:minimum-norm_point}) efficiently, when there are more domains.


\subsubsection{Scalability}
The graph convolutional layers in DMGE adopt the graph convolution operator defined in GCN~\cite{kipf2017semi}. However, GCN requires the full graph Laplacian, thus it is computationally expensive to apply GCN for large-scale graph embedding.

To apply DMGE for large-scale multi-graph embedding, we have the following strategies: 1) we can adopt GraphSAGE~\cite{hamilton2017inductive} as the graph convolutional layers in DMGE, as GraphSAGE generates embeddings by sampling and aggregating features from a node's local neighborhood, and only requires local graph structures; 2) we can replace the graph convolutional layers in DMGE with the graph attentional layers, which are presented in GAT~\cite{velivckovic2018graph}, as GAT is computationally efficient and parallelizable across all nodes in the graph, and doesn't require the entire graph structure upfront.

\section{Related Work}

\subsection{Embedding Methods}
Representation learning ~\cite{bengio2013representation} is one of the most fundamental problems in deep learning. As a practical application, effective embedding has been proved to be useful and achieve significant improvement in recommender systems (RSs) including: E-commerce~\cite{zhao2018learning,wang2018billion}, search ranking~\cite{chu2018deep,grbovic2018real} and social media~\cite{ying2018graph}.

The embedding methods in RSs can be divided into two categories: word embedding based methods and graph embedding based methods. The word embedding based methods~\cite{zhao2018learning,grbovic2018real} learn embedding by modeling the item co-occurrence in users' behavior sequences. Specifically, they model the items as words and user's behavior sequences as sentences, and apply word embedding methods~\cite{mikolov2013efficient,mikolov2013distributed} to represent items in a low-dimensional space. The graph embedding based methods~\cite{ying2018graph,wang2018billion} construct item graph based on users' behaviors, they model the items as nodes and item co-occurrences as edges, and apply the graph embedding methods~\cite{hamilton2017representation,cui2018survey,perozzi2014deepwalk,hamilton2017inductive} to learn embedding. However, these embedding methods are developed to learn embedding in a single domain, which fail to learn effective cross-domain embedding. Although there are several cross-domain recommendation methods~\cite{man2017cross}, they aim to improve the recommendation in target domain by transferring the information from source domain. In our work, we adopt graph neural network to learn more effective cross-domain embeddings to benefit all domains.


\subsection{Graph Neural Networks}

Graph neural networks (GNNs)~\cite{zhou2018graph,xu2019powerful} have emerged as a powerful approach for representation learning on graphs recently, such as GCN~\cite{kipf2017semi}, GraphSAGE~\cite{hamilton2017inductive}. Through a recursive neighborhood aggregation scheme, GNNs can generate node embedding by aggregating features of neighbors. In this part, we focus on reviewing related works about the convolution based GNNs, which can be categorized as spectral approaches and non-spectral approaches.

The spectral approaches depend on the theory of spectral graph convolutions. Bruna \emph{et al}.~\cite{bruna2014spectral} first propose a generalization of convolutional neural networks (CNNs) to graphs, however, it is computationally expensive. Defferrard \emph{et al}.~\cite{defferrard2016convolutional} design $K$-localized convolutional filters on graphs based on spectral graph theory, which is more computationally efficient. Kipf \emph{et al}.~\cite{kipf2017semi} limit the layer-wise convolution operation to $K=1$ to avoid overfitting, and propose the graph convolutional network (GCN), which can be applied to encode both local graph structure and features of nodes through layer-wise propagation. The non-spectral approaches operate spatial convolutions on the graph. Hamilton \emph{et al}.~\cite{hamilton2017inductive} propose GraphSAGE to generates node embeddings by sampling and aggregating features from a node's local neighborhood, which can be applied for large-scale graph embedding. However, these GNNs are developed for single graph embedding, which fail to learn effective multi-graph embedding.

\section{Conclusion}
In this paper, we focus on learning effective cross-domain representation. We propose the Deep Multi-Graph Embedding (DMGE) model, which is a multi-graph neural network based on multi-task learning. We construct the item graphs as a multi-graph based on users’ behaviors from different domains, and then design a graph neural network to learn multi-graph embedding in an unsupervised manner. Particularly, we introduce a multiple-gradient descent optimizer for efficiently training the model. We evaluate our approach on various large-scale real-world datasets, and the experimental results show that DMGE outperforms other state-of-the-art embedding methods in various tasks.

\section{Acknowledgments}
This work was partially supported by the National Key R\&D Program of China (2017YFB1001800) and the National Natural Science Foundation of China (No. 61772428, 61725205).

%


%
\bibliographystyle{ACM-Reference-Format}
\bibliography{sample-base}

%
\appendix

\end{document}